# Towards a Deep Learning Framework for Unconstrained Face Detection

Yutong Zheng[*]    Chenchen Zhu[*]    Khoa Luu    Chandrasekhar Bhagavatula
T. Hoang Ngan Le    Marios Savvides
CyLab Biometrics Center and the Department of Electrical and Computer Engineering,
Carnegie Mellon University, Pittsburgh, PA, USA
{yutongzh, chenchez, kluu, cbhagava, thihoanl}@andrew.cmu.edu, msavvid@ri.cmu.edu

## Abstract

*Robust face detection is one of the most important pre-processing steps to support facial expression analysis, facial landmarking, face recognition, pose estimation, building of 3D facial models, etc. Although this topic has been intensely studied for decades, it is still challenging due to numerous variants of face images in real-world scenarios. In this paper, we present a novel approach named Multiple Scale Faster Region-based Convolutional Neural Network (MS-FRCNN) to robustly detect human facial regions from images collected under various challenging conditions, e.g. large occlusions, extremely low resolutions, facial expressions, strong illumination variations, etc. The proposed approach is benchmarked on two challenging face detection databases, i.e. the Wider Face database and the Face Detection Dataset and Benchmark (FDDB), and compared against recent other face detection methods, e.g. Two-stage CNN, Multi-scale Cascade CNN, Faceness, Aggregate Chanel Features, HeadHunter, Multi-view Face Detection, Cascade CNN, etc. The experimental results show that our proposed approach consistently achieves highly competitive results with the state-of-the-art performance against other recent face detection methods.*

## 1. Introduction

Detection and analysis on human subjects using facial feature based biometrics for access control, surveillance systems and other security applications have gained popularity over the past few years. Several such biometrics systems are deployed in security checkpoints across the globe with more being deployed every day. Particularly, face recognition has been one of the most popular biometrics modalities attractive to security departments. Indeed, the uniqueness of facial features across individuals can be captured much more easily than other biometrics. In order for face recognition to take place, however, face detection usually needs to be done first. The problem of face detection has been intensely studied for decades with the aim of ensuring the generalization of robust algorithms to unseen face images [17, 24, 11, 25, 14]. Although the detection accuracy in recent face detection algorithms [2, 10, 19] has been highly improved, they are still far from achieving the same detection capabilities as a human due to a number of challenges in practice. For example, off-angle faces, large occlusions, low-resolutions and strong lighting conditions, as shown in Figure 1, are always the important factors that need to be considered.

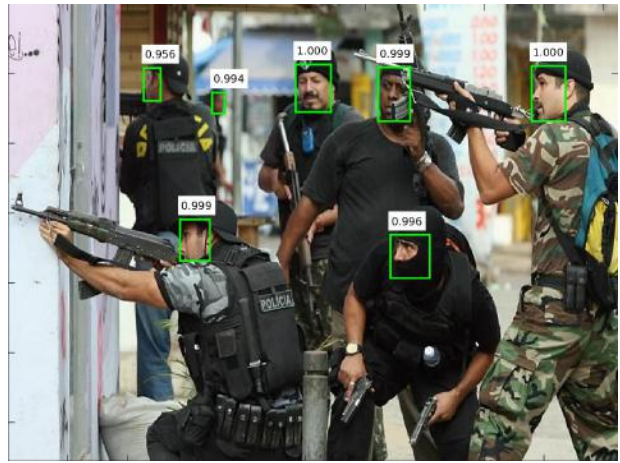

Figure 1. An example of face detection results using our proposed MS-FRCNN method. The proposed method can robustly detect faces across occlusions, facial expressions, poses, illumination and low resolution conditions on Wider Face database [21].

This paper presents an advanced Convolutional Neural Network (ConvNet) based approach named Multiple Scale Faster Region-based Convolutional Neural Network (MS-FRCNN) to handle the problem of face detection in digital face images collected under numerous challenging condi-tions, e.g. facial occlusions, strong illumination, off-angles, low-resolutions, etc. Our proposed approach extends the framework of the Faster R-CNN [15] by allowing it to span the receptive fields in the ConvNet in multiple deep feature maps. In other words, this process helps to synchronize both the global and the local context information for facial feature representation. Therefore, it is able to robustly deal with the challenges in the problem of robust face detection. Our proposed method introduces the Multiple Scale Regional Proposal Network (MS-RPN) to generate a set of region proposals and the Multiple Scale Region-based Convolutional Neural Network (MS-RNN) to extract the regions of interest (RoI) of facial regions. A confidence score is then computed for every RoI. Finally, the face detection system is able to decide the quality of the detection results by thresholding these generated confidence scores in given face images. The design of our proposed MS-FRCNN deep network for the problem of robust face detection can be seen in Figure 2.

---
[*]indicates equal contribution.

The proposed MS-FRCNN approach is evaluated on two challenging face detection databases and compared against numerous recent face detection methods. Firstly, the proposed MS-FRCNN method is compared against the standard Faster R-CNN method in the problem of face detection. It is evaluated on the Wider Face database [21], a large scale face detection benchmark dataset, to show its capability to detect face images in the wild, e.g. under occlusions, illumination, facial poses, low-resolution conditions, etc. It is also benchmarked on the Face Detection Data Set and Benchmark (FDDB) [7], a dataset of face regions designed for studying the problem of unconstrained face detection. The experimental results show that the proposed MS-FRCNN approach consistently achieves highly competitive results against the other state-of-the-art face detection methods. Finally, we present the limitations of the proposed MS-FRCNN method in the problem of face detection.

The rest of this paper is organized as follows. In section 2, we summarize prior work in face detection. Section 3 reviews a general deep learning framework, the background as well as the limitations of the Faster R-CNN in the problem of face detection. In Section 4, we introduce our proposed MS-FRCNN approach to the problem of robust face detection. Section 5 presents the experimental face detection results and comparisons obtained using our proposed approach on two challenging face detection databases, i.e. the Wider Face and the FDDB databases. Finally, our conclusions in this work are presented in Section 6.

## 2. Related Work

Face detection has been a well studied area of computer vision. One of the first well performing approaches to the problem was the Viola-Jones face detector [17]. It was ca-pable of performing real time face detection using a cascade of boosted simple Haar classifiers. The concepts of boosting and using simple features has been the basis for many different approaches [24] since the Viola-Jones face detector. These early detectors tended to work well on frontal face images but not very well on faces in different poses. As time has passed, many of these methods have been able to deal with off-angle face detection by utilizing multiple models for the various poses of the face. This increases the model size but does afford more practical uses of the methods. Some approaches have moved away from the idea of simple features but continued to use the boosted learning framework. Li and Zhang [11] used SURF cascades for general object detection but also showed good results on face detection.

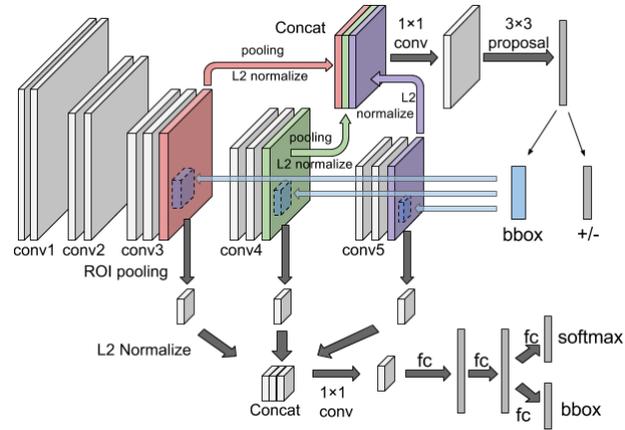

Figure 2. Our proposed Multiple Scale Faster R-CNN approach to unconstrained face detection.

More recent work on face detection has tended to focus on using different models such as a Deformable Parts Model (DPM) [25, 3]. Zhu and Ramanan's work was an interesting approach to the problem of face detection in that they combined the problems of face detection, pose estimation, and facial landmarking into one framework. By utilizing all three aspects in one framework, they were able to outperform the state-of-the-art at the time on real world images. Yu et al. [22] extended this work by incorporating group sparsity in learning which landmarks are the most salient for face detection as well as incorporating 3D models of the landmarks in order to deal with pose. Chen et al. [1] have combined ideas from both of these approaches by utilizing a cascade detection framework while simultaneously localizing features on the face for alignment of the detectors. Similarly, Ghiasi and Fowlkes [4] have been able to use heirarchical DPMs not only to achieve good face detection in the presence of occlusion but also landmark localization. However, Mathias et al. [14] were able to show that both DPM models and rigid template detectors similar to the Viola-Jones detector have a lot of potential that has not been adequately explored. By retraining these models with appropriately controlled training data, they were able to create face detectors that perform similarly to other, more complex state-of-the-art face detectors.

All of these approaches to face detection were based on selecting a feature extractor beforehand. However, there has been work done in using a ConvNet to learn which features are used to detect faces. Neural Networks have been around for a long time but have been experiencing a resurgence in popularity due to hardware improvements and new techniques resulting in the capability to train these networks on large amounts of training data. Li et al. [10] utilized a cascade of CNNs to perform face detection. The cascading networks allowed them to process different scales of faces at different levels of the cascade while also allowing for false positives from previous networks to be removed at later layers in a similar approach to other cascade detectors. Yang et al. [19] approached the problem from a different perspective more similar to a DPM approach. In their method, the face is broken into several facial parts such as hair, eyes, nose, mouth, and beard. By training a detector on each part and combining the score maps intelligently, they were able to achieve accurate face detection even under occlusions. Both of these methods require training several networks in order to achieve their high accuracy. Our method, on the other hand, can be trained as a single network, end-to-end, allowing for less annotation of training data needed while maintaining highly accurate face detection.

## 3. Background

The recent studies in deep ConvNets have achieved significant results in object detection, classification and modeling [9]. In this section, we review various well-known Deep ConvNets. Then, we show the current limitations of the Faster R-CNN, one of the state-of-the-art deep ConvNet methods in object detection, in the defined context of the face detection.

### 3.1. Deep Learning Framework

Convolutional Neural Networks are biologically inspired variants of multilayer perceptrons. The ConvNet method and its extensions, e.g. LeNet-5, HMAX, etc., imitate the characteristics of animal visual cortex systems that contain a complex arrangement of cells sensitive to receptive fields. In their models, the designed filters are considered as human visual cells in order to explore spatially local correlations in natural images. It efficiently presents the sparse connectivity and the shared weights since these kernel filters are replicated over the entire image with the same parameters in each layer. In addition, the pooling step, a form of down-sampling, plays a key role in ConvNet. Max-pooling is a popular pooling method for object detection and classification since max-pooling reduces computation for upper layers by eliminating non-maximal values and provides a small amount of translation invariance in each level.

Although ConvNets can explore deep features, they are very computationally expensive. The algorithm becomes more practical when implemented in a Graphics Processing Unit (GPU). The Caffe framework [8] is one of the fastest deep learning implementations using CUDA C++ for GPU computation. It also supports interfaces to Python/Numpy and MATLAB. It can be used as an off-the-shelf deployment of the state-of-the-art models. This framework is employed in our implementation.

### 3.2. Region-based Convolutional Neural Networks

One of the most important approaches in the object detection task is the family of Region-based Convolutional Neural Networks. The first generation of this family, R-CNN [6], applies the high-capacity deep ConvNet to classify given bottom-up region proposals. Due to the lack of labeled training data, it adopts a strategy of supervised pre-training for an auxiliary task followed by domain-specific fine-tuning. Then the ConvNet is used as a feature extractor and the system is further trained for object detection with Support Vector Machines (SVM). Finally, it performs bounding-box regression. The method achieves high accuracy but is very time-consuming. The system takes a long time to generate region proposals, extract features from each image, and store these features in a hard disk, which also takes up a large amount of space. At testing time, the detection process takes 47s per one image using VGG-16 network [16] implemented in GPU due to the slowness of feature extraction.

R-CNN [6] is slow because it processes each object proposal independently without sharing computation. Fast R-CNN [5] solves this problem by sharing the features between proposals. The network is designed to only compute a feature map once per image in a fully convolutional style, and to use ROI-pooling to dynamically sample features from the feature map for each object proposal. The network also adopts a multi-task loss, i.e. classification loss and bounding-box regression loss. Based on the two improvements, the framework is trained end-to-end. The processing time for each image significantly reduced to 0.3s.

Fast R-CNN accelerates the detection network using the ROI-pooling layer. However the region proposal step is designed out of the network hence still remains a bottleneck, which results in sub-optimal solution and dependence on the external region proposal methods. Faster R-CNN [15] addresses this problem by introducing the Region Proposal Network (RPN). A RPN is implemented in a fully convolutional style to predict the object bounding boxes and the objectness scores. In addition, the anchors are defined with different scales and ratios to achieve the translation invariance. The RPN shares the full-image convolution features with the detection network. Therefore the whole system is able to complete both proposal generation and detection computation within 0.2 seconds using very deep VGG-16 model [16]. With a smaller ZF model [23], it can reach the level of real-time processing.

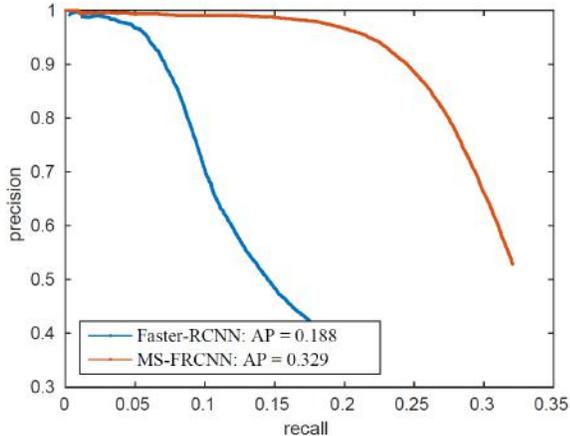

Figure 3. The face detection comparison between our proposed MS-FRCNN and the Faster R-CNN on the Wider Face validation set [21].

## 3.3. Limitations of Faster R-CNN

The Region-based CNN family, e.g. Faster R-CNN [15] and its variants [5], achieves the state-of-the-art performance results in object detection on the PASCAL VOC dataset. These methods can detect objects such as vehicles, animals, people, chairs, and etc. with very high accuracy. In general, the defined objects often occupy the majority of a given image. However, when these methods are tested on the challenging Microsoft COCO dataset [12], the performance drops a lot, since images contain more small, occluded and incomplete objects. Similar situations happen in the problem of face detection. We focus on detecting only facial regions that are sometimes small, heavily occluded and of low resolution (as shown in Figure 1). The detection network in designed Faster R-CNN is unable to robustly detect such tiny faces. The intuition point is that the Regions of Interest pooling layer, i.e. ROI-pooling layer, builds features only from the last single high level feature map. For example, the global stride of the 'conv5' layer in the VGG-16 model [16] is 16. Therefore, given a facial region with the sizes less than $16 \times 16$ pixels in an image, the projected ROI-pooling region for that location will be less than 1 pixel in the 'conv5' layer, even if the proposed region is correct. Thus, the detector will have much difficulty to predict the object class and the bounding box location based on information from only one pixel.

## 4. Our Approach to Robust Face Detection

This section presents our proposed Multiple Scale Faster R-CNN approach to robustly detect facial regions. Our approach utilizes the deep features encoded in both the global and the local representation for facial regions. Since the values of the filter responses range in different scales in each layer, i.e. the deeper a layer is, the smaller values of the filter responses are, there is a need for a further calibration process to synchronize the values received from multiple filter responses. The average feature for layers in Faster-RCNN are employed to augment features at each location.

### 4.1. Deep Network Architecture

In problem of face detection, the sizes of human faces in observed images are usually collected under low-resolutions, large occlusions and strong lighting conditions. It is an difficult task for the standard Faster R-CNN to robustly detect these facial regions. It is because the receptive fields in the last convolution layer (conv5) in the standard Faster R-CNN is quite large. For example, given a facial ROI region of sizes of $64 \times 64$ pixels in an image, its output in conv5 only contains $4 \times 4$ pixels, which is insufficient to encode informative features. When the convolution layers go deeper, each pixel in the corresponding feature map gather more convolutional information outside the ROI region. Thus, it contains higher proportion of information outside the ROI region if the ROI is very small. The two problems together, make the feature map of the last convolution layer less representative for small ROI regions.

Therefore, a combination of both global and local features, i.e. multiple scales, to enhance the global and local information in the Faster R-CNN model can help robustly detect facial regions. In order to enhance this capability of the network, we incorporate feature maps from shallower convolution feature maps, i.e. conv3 and conv4, to the convolution feature map conv5 for ROI pooling. Therefore, the network can robustly detect lower level facial features containing higher proportion of information in ROI regions.

Particularly, the defined network includes 13 convolution layers initialized using the pre-trained VGG-16 model. Right after each convolution layer, there is a ReLU layer. But only 4 of these layers are followed with pooling layers that shrink the spatial scale. Therefore the convolution layers are divided into 5 major parts, i.e. conv1, conv2, conv3, conv4 and conv5. Each contains 2 or 3 convolution layers, e.g. conv5_3. All of the convolution layers are shared between the MS-RPN and the MS-RNN, similar to the standard one [15]. When there are three convolution layers, i.e. conv3_3, conv4_3 and conv5_3, of each network, their outputs are also used as the inputs to three corresponding ROI pooling layers and normalization layers as shown in Figure 2. These L-2 normalization outputs are concatenated and shrunk to use as the inputs for the next network layers.

### 4.2. Multiple Scale Normalization

In our deep network architecture, features extracted from different convolution layers cannot be simply concatenated [13]. It is because the overall differences of the numbers of channels, scales of values and norms of feature map pixels among these layers. The detailed research shows that the deeper layers often contain smaller values than the shallower layers. Therefore, the larger values will dominate the smaller ones, making the system rely too much on shallower features rather than a combination of multiple scale features causing the system to no longer be robust.

In order to solve this problem, we introduce a normalization layer to the CNN architecture [13]. The system takes the multiple scale features and apply L2 normalization along the channel axis of each feature map. Then, since the channel size is different among layers, the normalized feature map from each layer needed to be re-weighted, so that their values are at the same scale. After that, the feature maps are concatenated to one single feature map tensor. This modification helps to stabilize the system and increase the accuracy. Finally, the channel size of the concatenated feature map is shrunk to fit right in the original architecture for the downstream fully-connected layers.

### 4.3. Deep Network Implementation

Before normalization, all feature maps are synchronized to the same size so that the concatenation can be applied. In the RPN, shallower feature maps are followed by pooling layers with certain stride to perform down-sampling. In the detection network, the ROI pooling layers already ensure that the pooled feature maps are at the same size. The implementation of L2 normalization layer follows the layer definition in [13], i.e. the system updates the re-weighting factor for each feature map during training. In our architecture, we combine feature maps from three layers, i.e. conv3, conv4 and conv5, of the convolution layers. They are normalized independently, re-weighted and concatenated. The initial value for the re-weighting factor needs to be set carefully to make sure the downstream values are at reasonable scales when training is initialized.

Additionally, in order to shrink the channel size of the concatenated feature map, a $1 \times 1$ convolution layer is then employed. Therefore the channel size of final feature map is at the same size as the original fifth convolution layer in Faster-RCNN, as shown in Figure 2.

## 5. Experimental Results

This section presents the face detection bechmarking in our proposed MS-FRCNN approach on the Wider Face database [21] and the Face Detection Data Set and Benchmark (FDDB) [7]. In Section 5.1, we present the training steps on the Wider Face database. In Section 5.2, the face detection results using MS-FRCNN and Faster R-CNN are compared on the Wider Face database. Section 5.3 evaluates the proposed MS-FRCNN against other recently published face detection methods on the Wider Face database. In Section 5.4, our MS-FRCNN is also evaluated on the challenging FDDB face database. Finally, we analyze some cases when MS-FRCNN fails in detecting human faces.

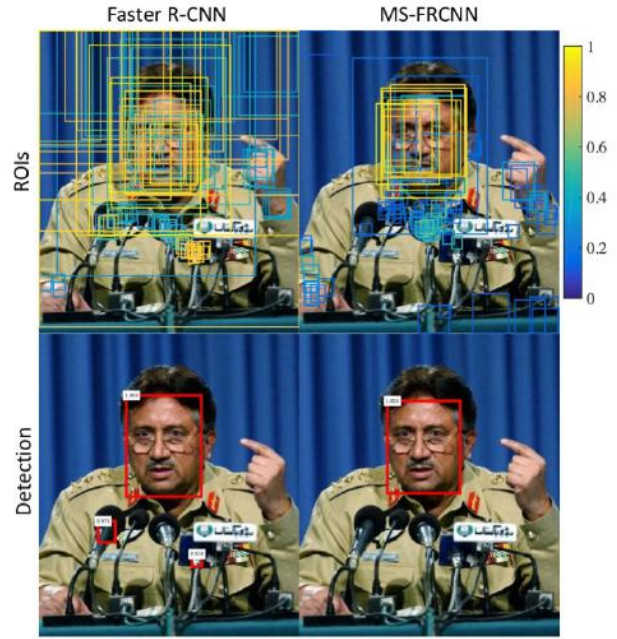

Figure 4. The proposed ROI candidates (upper) and the detection results (lower) using Faster R-CNN (left) and and MS-FRCNN (right). The color bar shows the confidence of each ROI region given by RPN.

### 5.1. Training Data

The Wider Face is a public face detection benchmark dataset. It contains 393,703 labeled human faces from 32,203 images collected based on 61 event classes from internet. The database has many human faces with a high degree of pose variation, large occlusions, low-resolutions and strong lighting conditions. The images in this database are organized and split into three subsets, i.e. training, validation and testing. Each contains 40%, 10% and 50% respectively of the original databases. The images and the ground-truth labels of the training and the validation sets are available online for experiments. However, in the testing set, only the testing images (not the ground-truth labels) are available online. All detection results are sent to the database server for evaluating and receiving the Precision-Recall curves.

In our experiments, the proposed MS-FRCNN is trained on the training set of the Wider Face dataset containing

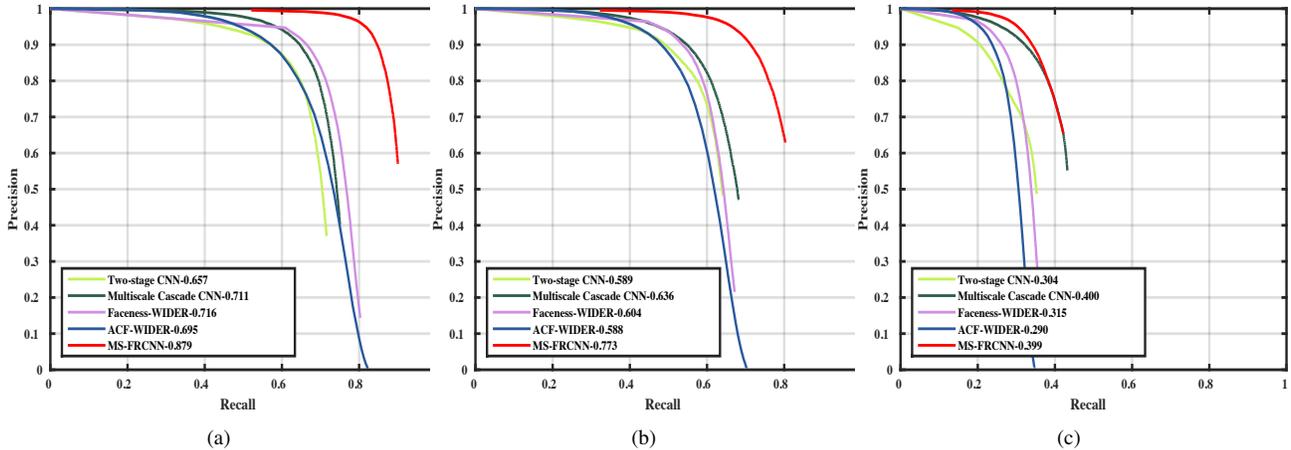

Figure 5. Precision-Recall curves obtained by our proposed MS-FRCNN (red) and the other methods, i.e. Two-stage CNN [21], Multi-scale Cascade CNN [21], Faceness [20], and Aggregate Channel Features (ACF) [18]. All methods trained and tested on the same training and testing set on the Wider Face database. (a): easy level, (b): medium level and (c): hard level. Our method achieves the state-of-the-art results with the highest AP values of 0.879 (easy), 0.773 (medium) and 0.399 (hard) among the methods on this database.

159,424 annotated faces collected in 12,880 images. The trained model on this database are used in testing in the following experiments.

## 5.2. MS-FRCNN v.s. Faster R-CNN in Face Detection

In this experiment, we compare the face detection capabilities using our proposed MS-FRCNN approach and the Faster R-CNN. Both methods are trained on the same training set as shown in Section 5.1. In addition, both methods perform under the same conditions. The Precision-Recall curves for the MS-FRCNN and the Faster R-CNN in the problem of face detection are shown in Figure 3. As shown in this figure, the proposed MS-FRCNN method strongly outperforms the Faster R-CNN in the problem of face detection in both the precision and the recall values. Our method achieves the Average Precision (AP) value of 0.329 while the Faster R-CNN has the AP value of 0.188. This experiment shows that the MS-FRCNN provides a more appropriate deep model to solve the problem of unconstrained face detection under various challenging conditions.

Figure 4 shows an example where the proposed MS-FRCNN consistently generates ROIs among a human face candidate while the Faster R-CNN has many ROIs confusing the classifier.

## 5.3. Face Detection on Wider Face Database

In this experiment, the training phase is the same as in Section 5.1. During the testing phase, the face images in the testing set are divided into three parts based on their detection rates on EdgeBox [26]. In other words, face images are divided into three levels according to the difficulties of the detection, i.e. Easy, Medium and Hard [21]. The proposed MS-FRCNN method is compared against recent face detection methods, i.e. two-stage CNN [21], Multiscale Cascade CNN [21], Faceness [20], and Aggregate channel features (ACF) [18]. All these methods are trained on the same training set and tested on the same testing set. The Precision-Recall curves and AP values are shown in Figure 5. Our method has highly competitive results with the state-of-the-art performance against recent face detection methods. It achieves the best average precision in all level faces, i.e. AP = 0.879 (easy), 0.773 (medium) and 0.399 (hard). Figure 6 shows some examples of face detection results using the proposed MS-FRCNN on this database.

## 5.4. Face Detection on FDDB database

To show that our method generalizes well to other standard datasets, the proposed MS-FRCNN is also benchmarked on the FDDB database. It is a standard database for testing and evaluation of face detection algorithms. It contains annotations for 5,171 faces in a set of 2,845 images taken from the Faces in the Wild dataset. We use the same model trained on Wider Face database presented in Section 5.1 to perform the evaluation on the FDDB database.

The evaluation is performed based on the discrete criterion, i.e. if the ratio of the intersection of a detected region with an annotated face region is greater than 0.5, it is considered as a true positive detection. The evaluation is proceeded following the FDDB evaluation protocol and compared against the published methods provided in the protocol. The proposed MS-FRCNN approach outperforms most of the published face detection methods and achieves a very high recall rate comparing against all other methods (as shown Figure 7). This is concrete evidence to demonstrate that MS-FRCNN robustly detects unconstrained faces. Fig-ure 8 shows some examples of the face detection results using the proposed MS-FRCNN on the FDDB database.

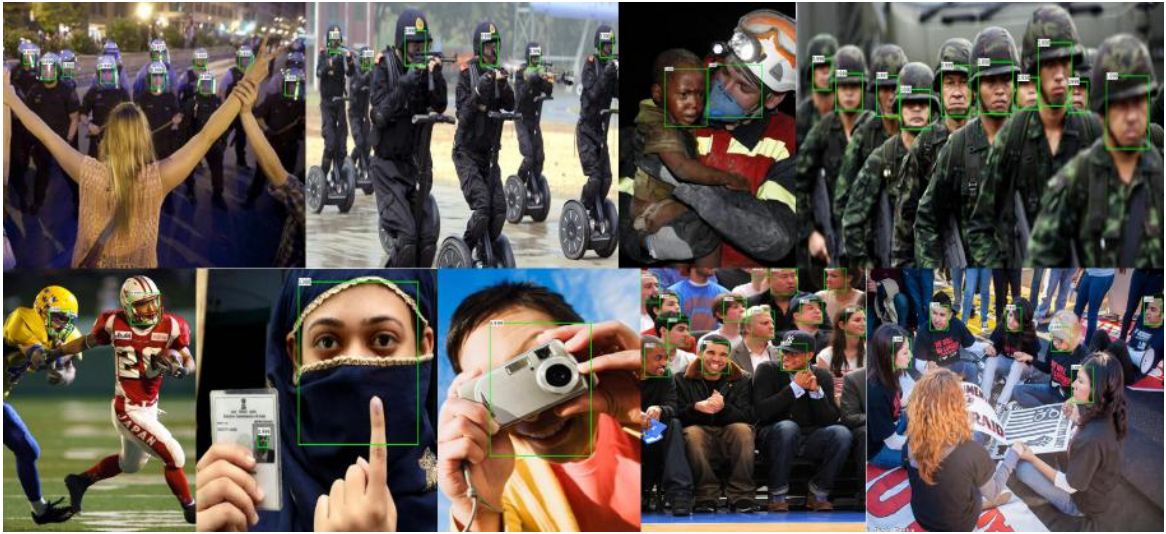

Figure 6. Some examples of face detection results using our proposed MS-FRCNN method on Wider Face database [21].

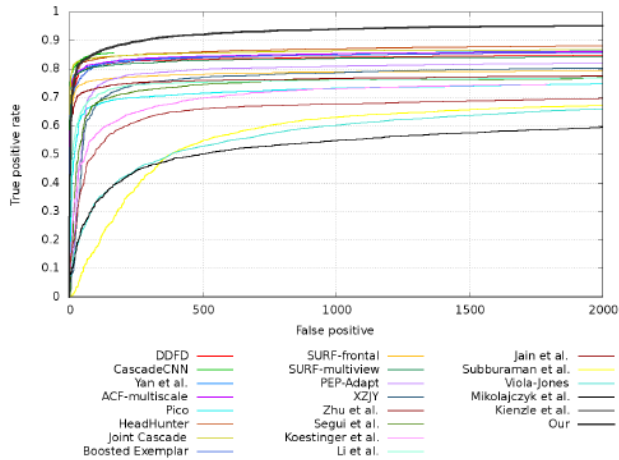

Figure 7. ROC curves on our proposed MS-FRCNN and the other published methods on FDDB database. Our method achieves the best recall rate on this database.

## 5.5. When MS-FRCNN Fails in Face Detection

In Wider Face database, there are many tiny labeled facial regions that need to be learned. The proposed method is trained on too many faces in those low quality conditions. Indeed, the human facial features in those facial regions are very limited. Therefore, the algorithm over-fits deep facial features in some cases. Given a new testing image, the trained system may mislabel some small regions with complicated patterns as human faces as shown in Figure 9. This is the point we will explore for a better solution in future.

## 6. Conclusion

This paper has presented our proposed MS-FRCNN approach to robustly detect human facial regions from images collected under various challenging conditions, e.g. highly occlusions, low resolutions, facial expressions, illumination variations, etc. The approach is benchmarked on two challenging face detection databases, i.e. the Wider Face database and the FDDB, and compared against recent other face detection methods, e.g. Two-stage CNN, Multiscale Cascade CNN, Faceness, Aggregate Chanel Features (ACF), etc. The experimental results show that our proposed approach consistently achieves very competitive results against the-state-of-the-art methods.

## References

[1] D. Chen, S. Ren, Y. Wei, X. Cao, and J. Sun. Joint cascade face detection and alignment. In *ECCV*, volume 8694, pages 109–122. 2014.

[2] S. S. Farfade, M. J. Saberian, and L.-J. Li. Multi-view face detection using deep convolutional neural networks. In *ICMR*, pages 643–650, 2015.

[3] P. Felzenszwalb, R. Girshick, D. McAllester, and D. Ramanan. Object detection with discriminatively trained part-based models. *IEEE Trans. on PAMI*, 32(9):1627–1645, Sept 2010.

[4] G. Ghiasi and C. Fowlkes. Occlusion coherence: Localizing occluded faces with a hierarchical deformable part model. In *CVPR*, 2014.

[5] R. Girshick. Fast r-cnn. In *ICCV*, pages 1440–1448, 2015.

[6] R. Girshick, J. Donahue, and J. M. T. Darrell. Region-based convolutional networks for accurate object detection and semantic segmentation. *IEEE Trans. on PAMI*, 2015.

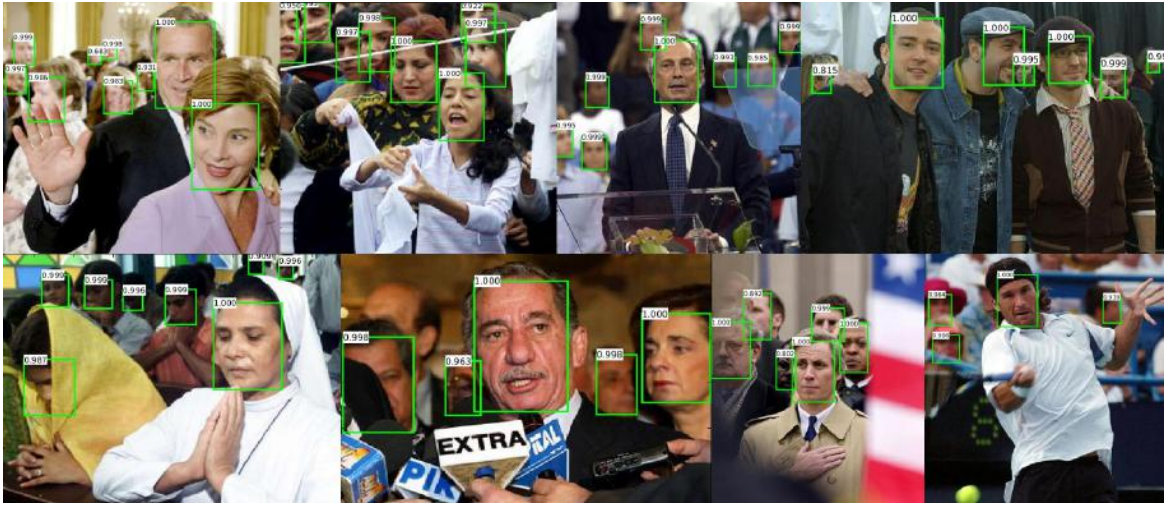

Figure 8. Some examples of face detection results using our proposed MS-FRCNN method on FDDB database [7].

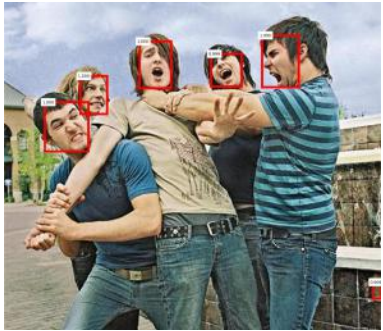

Figure 9. An example of face detection with a failure using MS-FRCNN on Wider Face database [21]. The wrong face region is shown in a tiny window on the right hand side of the image.

[7] V. Jain and E. Learned-Miller. Fddb: A benchmark for face detection in unconstrained settings. Technical Report UM-CS-2010-009, University of Massachusetts, Amherst, 2010.

[8] Y. Jia, E. Shelhamer, J. Donahue, S. Karayev, J. Long, R. Girshick, S. Guadarrama, and T. Darrell. Caffe: Convolutional architecture for fast feature embedding. *arXiv preprint arXiv:1408.5093*, 2014.

[9] A. Krizhevsky, I. Sutskever, and G. E. Hinton. Imagenet classification with deep convolutional neural networks. In *NIPS*, pages 1097–1105. 2012.

[10] H. Li, Z. Lin, X. Shen, J. Brandt, and G. Hua. A convolutional neural network cascade for face detection. In *CVPR*, June 2015.

[11] J. Li and Y. Zhang. Learning surf cascade for fast and accurate object detection. In *CVPR*, pages 3468–3475, June 2013.

[12] T.-Y. Lin, M. Maire, S. Belongie, J. Hays, P. Perona, D. Ramanan, P. Dollár, and C. L. Zitnick. Microsoft coco: Common objects in context. In *ECCV*, pages 740–755. 2014.

[13] W. Liu, A. Rabinovich, and A. C. Berg. Parsenet: Looking wider to see better. *arXiv preprint arXiv:1506.04579*, 2015.

[14] M. Mathias, R. Benenson, M. Pedersoli, and L. Van Gool. Face detection without bells and whistles. In *ECCV*, volume 8692, pages 720–735. 2014.

[15] S. Ren, K. He, R. B. Girshick, and J. Sun. Faster R-CNN: towards real-time object detection with region proposal networks. *CoRR*, abs/1506.01497, 2015.

[16] K. Simonyan and A. Zisserman. Very deep convolutional networks for large-scale image recognition. *arXiv preprint arXiv:1409.1556*, 2014.

[17] P. Viola and M. Jones. Robust real-time face detection. *IJCV*, 57:137–154, 2004.

[18] B. Yang, J. Yan, Z. Lei, and S. Z. Li. Aggregate channel features for multi-view face detection. In *IJCB*, pages 1–8. IEEE, 2014.

[19] S. Yang, P. Luo, C.-C. Loy, and X. Tang. From facial parts responses to face detection: A deep learning approach. In *ICCV*, Dec. 2015.

[20] S. Yang, P. Luo, C.-C. Loy, and X. Tang. From facial parts responses to face detection: A deep learning approach. In *ICCV*, pages 3676–3684, 2015.

[21] S. Yang, P. Luo, C. C. Loy, and X. Tang. Wider face: A face detection benchmark. In *The IEEE Conference on Computer Vision and Pattern Recognition (CVPR)*, pages 5525–5533. IEEE, 2016.

[22] X. Yu, J. Huang, S. Zhang, W. Yan, and D. Metaxas. Pose-free facial landmark fitting via optimized part mixtures and cascaded deformable shape model. In *ICCV*, pages 1944–1951, Dec 2013.

[23] M. D. Zeiler and R. Fergus. Visualizing and understanding convolutional networks. In *ECCV*, pages 818–833. 2014.

[24] C. Zhang and Z. Zhang. A survey of recent advances in face detection. Technical Report MSR-TR-2010-66, June 2010.

[25] X. Zhu and D. Ramanan. Face detection, pose estimation, and landmark localization in the wild. In *CVPR*, pages 2879–2886, June 2012.

[26] C. L. Zitnick and P. Dollár. Edge boxes: Locating object proposals from edges. In *ECCV*, pages 391–405. Springer, 2014.